\documentclass{article}
\usepackage{spconf,amsmath,graphicx}

\usepackage{enumitem}
\setlist{nosep, leftmargin=14pt}
\usepackage{cite}
\usepackage{amsmath,amssymb,amsfonts}
\usepackage{algorithm}
\usepackage{algorithmic}
\usepackage{graphicx}
\usepackage{textcomp}
\usepackage{xcolor}

\usepackage{booktabs}
\usepackage[table]{xcolor}
\definecolor{Gray}{gray}{0.9}
\usepackage{multirow}
\usepackage{array}

\usepackage{graphicx}
\usepackage{subfig} 

\usepackage{booktabs}
\usepackage[table]{xcolor}
\usepackage{graphicx}
\usepackage{subcaption}  
\usepackage{pgfplots}
\pgfplotsset{compat=1.18}
\usepackage{amsmath,amssymb}

\usepackage[colorlinks=true,
            linkcolor=blue,
            citecolor=blue,
            urlcolor=blue]{hyperref}

\usepackage{tikz}
\usepackage{pgfplots}
\pgfplotsset{compat=1.18}

\usepackage{amsmath,amssymb}

\def\*#1{\mathbf{#1}}
\def\R{\mathbb{R}}
\def\-#1{\mathbb{#1}}

\def\BibTeX{{\rm B\kern-.05em{\sc i\kern-.025em b}\kern-.08em
    T\kern-.1667em\lower.7ex\hbox{E}\kern-.125emX}}

\title{Dual-Domain Equivariant Generative Adversarial Network for Multimodal CT-PET Synthesis}

\name{Gabriel~Steele$^{\star}$ \qquad Alzahra~Altalib$^{\star \dagger}$ \qquad Alessandro Perelli$^{\ddagger}$}

\address{$^{\star}$ School of Science and Engineering, University of Dundee, DD1 4HN, UK \\ $^{\dagger}$ Faculty of Applied Medical Sciences, Jordan University of Science and Technology, 21410, Jordan \\ $^{\ddagger}$ School of Cardiovascular and Metabolic Health, University of Glasgow, G12 8TA, UK}

\begin{document}

\maketitle

\begin{abstract}
We present a Dual-Domain Equivariant Generative Adversarial Network (DDE-GAN) for multimodal CT-PET image synthesis. Traditional GAN-based approaches often operate solely in the spatial domain and ignore geometric consistency, resulting in limited structural fidelity. DDE-GAN addresses these challenges by jointly learning from both spatial and frequency (Fourier) domains, capturing complementary anatomical and spectral information. Furthermore, rotational equivariance embedded in the physics of the CT and PET measurements are integrated into the loss of both the generator and discriminator to ensure consistent responses under rotations, improving anatomical accuracy. A hierarchical dual-domain training strategy enforces intra- and inter-domain consistency through multi-stage loss functions. Evaluated on the HECKTOR 2022 CT-PET dataset, DDE-GAN achieves superior synthesis quality over baseline models for CT-PET image synthesis. The results demonstrate that combining dual-domain learning with geometric equivariance substantially enhances multimodal image synthesis accuracy and robustness, enabling practical applications in PET completion and data augmentation.
\end{abstract}


\section{Introduction}
	
Multimodal imaging, particularly Computed Tomography-Positron Emission Tomography (CT-PET), has become indispensable in modern clinical workflows, providing complementary metabolic and anatomical information essential for diagnosis, staging, and treatment planning in oncology, neurology, and cardiology \cite{Townsend2008}. Despite its diagnostic power, dual-modality acquisition is associated with increased cost, scan time, and patient radiation exposure. Moreover, in many real-world scenarios, including low-resource settings, retrospective studies, or emergency situations as one of the modalities may be missing, corrupted, or misaligned. This motivates the growing interest in cross-modality image synthesis, where one modality like CT is used to generate the other like PET \cite{Xiang2018}, \cite{Nie2017}. However, it is of outmost importance to ensure consistency across different measurements' modalities and furthermore consistent outputs under transformations like rotation or translation undermines their generalizability and clinical reliability \cite{Cohen2016}, \cite{Weiler2018}. Early methods for modality synthesis relied on registration-based or patch-based approaches, which were limited by their reliance on handcrafted features and spatial alignment. 

Recent advances in deep generative models, especially Generative Adversarial Networks (GANs), have enabled promising results in medical image-to-image translation tasks \cite{Isola2017}. In the medical domain, GANs have been widely adopted for MRI-to-CT \cite{Wolterink2017}, CT-to-PET \cite{Xiang2018}, and other synthesis tasks. In \cite{Nie2017} one of the earliest deep learning-based frameworks for cross-modality image synthesis was introduced, leveraging contextual information with a conditional GAN. A modality-invariant latent representations was proposed in \cite{Chartsias2018} to improve the generalizability of synthetic images. 
More recent work, such as \cite{Huang2021}, utilized 3D convolutional GANs to synthesize PET from CT, improving realism and metabolic plausibility. Nevertheless, existing approaches face key challenges. First, most models operate solely in the image or latent space, failing to fully leverage the complementary spectral characteristics of PET and CT data. Second, these models often lack mechanisms to preserve geometric consistency under spatial transformations, leading to anatomically implausible hallucinations. However, most existing GAN-based models operate solely in the spatial domain, limiting their ability to capture global or frequency-sensitive features relevant to multimodal synthesis. Dual-domain learning has gained traction as a means to leverage complementary information from different representations. In medical imaging, some studies have explored frequency-aware architectures for denoising or reconstruction, but dual-domain learning for modality synthesis remains under-explored. Our method builds on this idea by learning from both image and frequency domains, allowing the network to capture detailed anatomical structures and broader contextual patterns. Most deep networks are not inherently equivariant to transformations such as rotation, translation, or reflection, which can be problematic in medical applications where orientation and geometry are clinically relevant. Group-equivariant convolutional neural networks (G-CNNs) enforce symmetry constraints, improving generalization and geometric consistency which have been applied for 3D imaging \cite{Weiler2018}, demonstrating benefits in segmentation and classification. An alternative way to enforce equivariance consistency is to exploits the group invariance present in PET-CT distributions directly from measurement data \cite{chen2022}. Yet, equivariance has not been widely incorporated into generative models for modality synthesis, despite its potential to improve anatomical plausibility. Our work addresses this gap by integrating equivariant loss constraints into both the generator and discriminator architectures of a GAN framework.
	
\subsection{Main Contribution}
To address these limitations, we propose a novel framework: the Dual-Domain Equivariant Generative Adversarial Network (DDE-GAN) for multimodal PET-CT synthesis. Our method introduces two major innovations a) Dual-domain learning: We jointly process image data in both the spatial (image) domain and the frequency (Fourier) domain, allowing the model to learn high-frequency anatomical detail as well as low-frequency contextual information. This is inspired by the growing use of frequency-aware models in medical image reconstruction and denoising tasks; b) Equivariant design: We incorporate robust equivariant imaging to enforce this property in the dual domain loss function into both the generator and discriminator to preserve geometric consistency. By ensuring that transformations (e.g., rotation, reflection) in the input are reflected in the output, our model maintains anatomical plausibility and improves robustness to variations in patient orientation. To our knowledge, equivariant architectures have not yet been applied to generative modelling for modality synthesis. We propose the first Dual-Domain Equivariant GAN (DDE-GAN) for CT-PET synthesis, jointly learning from both spatial and frequency representations to enhance cross-modality translation quality and the geometric equivariance constraints are embedded into the generative training loss, improving structural fidelity and generalization under spatial transformations. We demonstrate significant improvements over state-of-the-art baselines on two public datasets, showing enhanced anatomical accuracy, synthesis realism, and modality consistency.
	
Our findings suggest that DDE-GAN has practical applications in PET generation from low-dose CT, and data augmentation for multimodal training pipelines.

\section{Dual-Domain Equivariant GAN}
	
Generalised Dual-Domain Generative Framework \cite{zhang2023} for reconstruction makes use of information in both the imaging and raw data domains to generate additional cost functions that introduce different consistency constraints allowing for incremental stages of model training.
We define $\*x_s\in\R^n$ the source image and $\*x_s\in\R^n$ the target image that can be either considered as PET or CT and we indicate $\*F\in\R^{m\times n}$ as the linear operator representing the ray-tracing discretization physical attenuation model, such that $\*y_{s,t}=\*F_{s,t}\*x_{s,t}$. This procedure employs hierarchical bi-directional consistency constraints to train the networks in three stages. The first of these consistency constraints is intra-domain consistency $S_1$. Different networks $G_t^I, G_s^I$ operating in the image domain and $G_t^A, G_s^A$ in the measurement domain, are trained to construct target images and inversely construct the source images from target data for both domains. Initially the networks are trained individually and the $L_1$ norm cost functions in Eqs. (\ref{eq:gddgf_S1_loss}) are computed in the training in stage $S_1$ 
\begin{eqnarray}\label{eq:gddgf_S1_loss}
	L_t^I &=& \-E_{\*x_s, \*x_t} \left\|G_t^I(\*x_s) - \*x_t       \right\|_1 \\
	L_s^I &=& \-E_{\*x_s, \*x_t} \left\|G_s^I(\*x_t) - \*x_s       \right\|_1 \nonumber\\
	L_t^A &=& \-E_{\*x_s, \*x_t} \left\|G_t^A(\*F_s(\*x_s)) - \*F_t(\*x_t) \right\|_1 \nonumber\\
	L_s^A &=& \-E_{\*x_s, \*x_t} \left\|G_s^A(\*F_t(\*x_t)) - \*F_s(\*x_s) \right\|_1 \nonumber
\end{eqnarray}
With the trained networks from the first stage, the second stage $S_2$ measures consistency constraint by introducing inter-domain elements. Networks are paired to compare generated images, through the forward operators $\*F_{s,t}$ between domains. Inter-domain stage $S_2$ cost functions 
\begin{eqnarray}\label{eq:gddgf_S2_loss}
	L_t^I &=& \-E_{\*x_s, \*x_t}      \left\| G_s^I(\*x_s) - \*x_t                   \right\|_1 \\
	& & + \; \lambda_1 \-E_{\*x_s} \left\| G_t^I(\*x_s) - \*F_t^{-1}(G_t^A(\*F_s(\*x_s))) \right\|_1 \nonumber \\
	L_t^A &=& \-E_{\*x_s, \*x_t}      \left\| G_t^A(\*F_s(\*x_s)) - \*F_t(\*x_t) \right\|_1  \nonumber \\
	& & + \; \lambda_2 \-E_{\*x_s} \left\| G_t^A(\*F_s(\*x_s)) - \*F_t(G_t^I(\*x_s)) \right\|_1 \nonumber \\
		L_s^I &=& \-E_{\*x_s, \*x_t}      \left\| G_s^I(\*x_t) - \*x_s                   \right\|_1  \nonumber \\
	& & \; + \lambda_3 \-E_{\*x_s} \left\| G_t^I(\*x_t) - \*F_s^{-1}(G_s^A(\*F_t(\*x_t))) \right\|_1 \nonumber \\
		L_s^A &=& \-E_{\*x_s, \*x_t}      \left\| G_s^A(\*F_t(\*x_t)) - \*F_s(\*x_s) \right\|_1 \nonumber \\
	& & \; + \lambda_4 \-E_{\*x_s} \left\| G_s^A(\*F_t(\*x_t)) - \*F_t(G_s^I(\*x_t)) \right\|_1 \nonumber
\end{eqnarray} 
build on the intra-domain consistency to compare the results of network reconstructions within different domains. $\lambda$ is a scalar value that changes the weighting of the terms to the loss value and $\*F^{-1}_{s,t}$ is the back-projection operator. 

\begin{figure}[!h]

\begin{minipage}[b]{\linewidth}
  \centering
  \centerline{\includegraphics[width=.9\textwidth]{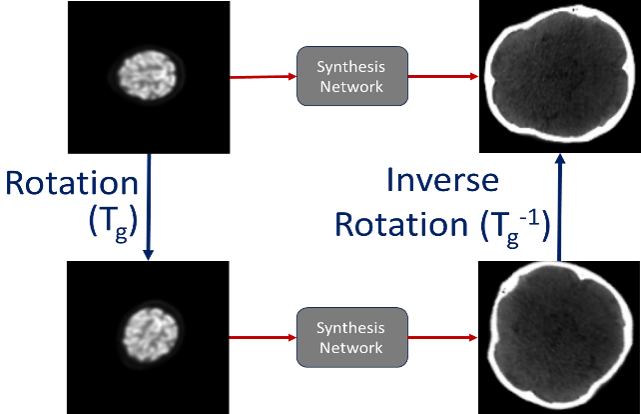}}
\end{minipage}
\caption{Rotational equivariance for multimodal PET/CT image synthesis.} \label{fig_equiv}
\end{figure}

By design both PET and CT inherited rotational equivariance by the physics of the acquisition, therefore the key idea is to enforce this constraint into the training loss. Defining $\*T_g: \-R^n\rightarrow \-R^n$ as a rotational operator, the equivariance constraint can be enforced by imposing 
\begin{equation}\label{eq:Trot}
    \*T_g\big( G^I_s \left( \*F_s(\*x_s) \right) \big) = G^I_s \big( \*F_s\left(\*T_g(\*x_s)\right) \big)
\end{equation}
and this can be visually depicted as in Fig. \ref{fig_equiv}.

\noindent Based on Eq. (\ref{eq:Trot}), we design he equivariant loss function for Stage $S_{3, eq}$ as 
\begin{eqnarray}\label{eq:s3_eq}
	L_t^I & = & \-E_{\*x_s, \*x_t} \left\| G_t^I(\*x_s) - \*x_s \right\|_1 \\
    & & + \;\xi_1 \-E_{\*x_s}\left\| G_s^I (G_t^I(x_s)) - \*x_s                         \right\|_1 + \nonumber \\
	&   & + \;\xi_2 \-E_{\*x_s} \left\|\*F^{-1}(G_s^A (\*F (G_t^I (\*T_g(\*x_s))))) - \*T_g(\*x_s) \right\|_1 \nonumber\\
	L^I_s &=& \-E_{\*x_s, \*x_t}    \left\| G_t^I(\*x_t) - \*x_s                      \right\|_1 \\
	& & + \;\xi_3 \-E_{\*x_t}   \left\| G_t^I( G_s^I(\*x_t)) - \*x_t              \right\|_1 + \nonumber \\ 
	& & + \;\xi_4 \-E_{\*x_t} \left\| \*F^{-1}\left( G_t^A( \*F (G_s^I(\*T_g(\*x_t))))\right) - \*T_g(\*x_t) \right\|_1 \nonumber
\end{eqnarray}
where the last term in Eq. (\ref{eq:s3_eq}) compares the error between the rotated source image $\*T_g(\*x_s)$ with the equivariant transformed $\*F^{-1}$ domain source image $G^A_s(\*F_t(G^I_t(\*T_g(\*x_s))))$. The input source image is rotated $\*T_g(\*x_s)$ is reconstructed to represent the target image $G^I_t(\*T_g(\*x_s))$, which is transformed from the image domain $F(G^I_t(\*T_g(\*x_s)))$ and reconstructed into the $F$ domain target image $G^A_s(F_t(G^I_t(\*x_s)))$. $\xi_4$ is a constant that scales the impact of the term to the loss value.

\section{Validation and Results}

\begin{figure}[!h]
\begin{minipage}[b]{1.0\linewidth}
  \centering
\centerline{\includegraphics[width=\textwidth]{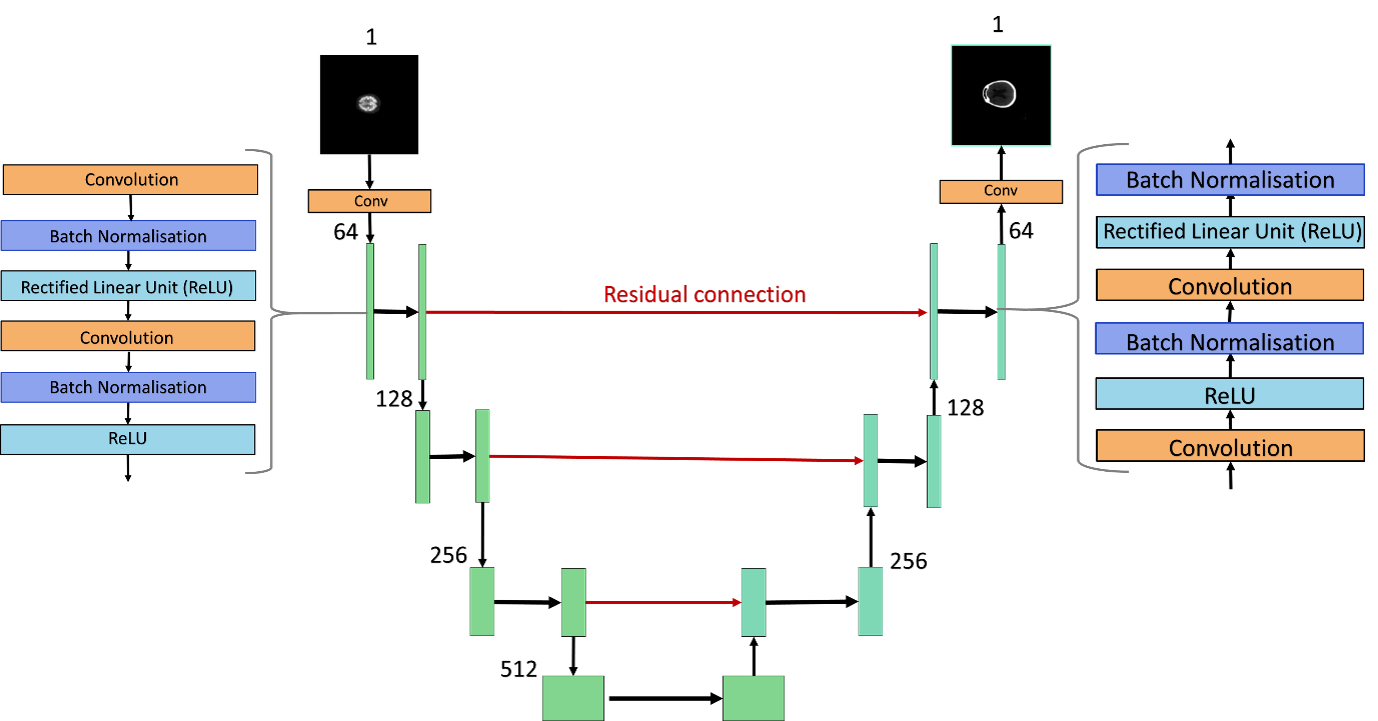}}
\end{minipage}
\caption{Residual U-Net network architecture in DDE-GAN.} \label{fig_ResUnet}
\end{figure}

The selected dataset for the study is sourced from the HEad and neCK TumOR segmentation in PET/CT images (HECKTOR) 2022 \cite{andrearczyk2022}. The datasets were composed of data from 10 different centres, each using different imaging procedures and machines. The training and testing datasets were used, resulting in the selection of CHUS, HMR and CHUV for training, and MDA for both training and testing. This selection resulted in a dataset of 341 PET and CT acquisitions, yielding an average total of $125,852$ images for training pre-processing and a set of 200 acquisitions, which corresponded to approximately $74,400$ images for testing pre-processing. These centres were selected to be included in the datasets because the images could be processed more effectively.
The refined dataset varied across several key features, notably in terms of resolution and the quantity of images per acquisition. Before processing, the PET image dataset was aligned and padded to $256\times 256$ pixels resolution. On the other hand, the CT image dataset had a mean resolution of $510\times 510$ pixels. The alignment between CT and PET images was performed by comparing the number of images yielded in each acquisition. The networks $G^{A, I}_{s,t}$ have been implemented using the U-Net architecture depicted in Fig. \ref{fig_ResUnet}.

\begin{figure}[!h]
	\begin{center}
		\small\addtolength{\tabcolsep}{-18pt}
		\renewcommand{\arraystretch}{0.1}
			\begin{tabular}{c}
				\small (a) CT ground \hspace{.3cm} \small (b) DD-GAN \hspace{.3cm} \small (c) DD-GAN  \hspace{.3cm} \small (d) DDE-GAN \\
                \hspace{.3cm} \small truth \hspace{1.3cm} \small \cite{zhang2023} \hspace{1.5cm} \small $S_2$  \hspace{2.3cm}  \\
				\begin{tikzpicture}
					\node {\includegraphics[width=.5\textwidth]{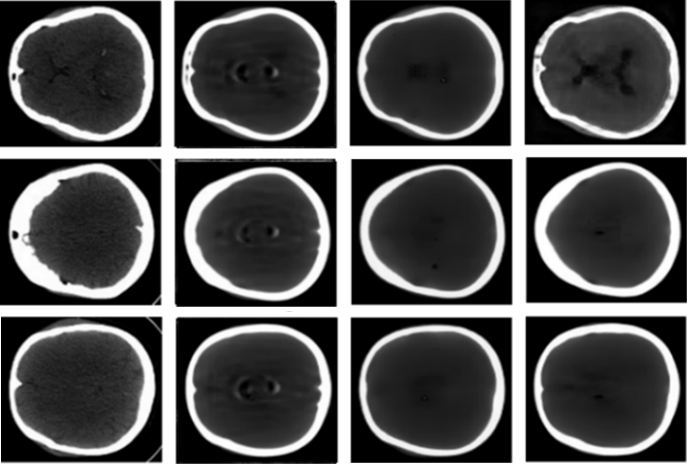}};
				\end{tikzpicture}
			\end{tabular}
		\caption{Synthesised PET images from CT using: b) DD-GAN \cite{zhang2023}, c) DD-GAN $S_2$ and d) proposed DDE-GAN.}\label{fig:PET_CT}
	\end{center}
\end{figure}

The qualitative results for the problem of synthesis of PET images from CT are shown in Fig. \ref{fig:PET_CT} where the synthesis images using CycleGAN \cite{zhang2023} and the proposed DD-GAN at different stages $S_1$ - $S_3$ are compared. It is possible to note how adding the equivariant loss at stage $S_3$, the details of the PET image are noticeably improved compared with previous stages and also with CycleGAN which is not fully converged given the same amount of training epochs. These observations are confirmed also by the quantitative results in terms of SSIM and PSNR summarised in Table \ref{tab:metrics} where the proposed DDE-GAN is consistently increased the PSNR of $\approx 6$ dB with reduced variability on the test set.

\begin{table}[!h]
\centering
\caption{Quantitative comparison of models on the test set for CT-PET image synthesis. Best scores are in bold.}
\label{tab:metrics}
\begin{tabular}{p{3cm}p{2cm}p{2cm}}
\hline
\rowcolor{Gray}\textbf{Method} & \textbf{SSIM} & \textbf{PSNR (dB)} \\ 
\hline
DD-GAN \cite{zhang2023} & $0.75 \pm 0.18$ & $25.42 \pm 2.35$ \\   \hline
DD-GAN $S_2$ & $0.80 \pm 0.11$ & $21.51 \pm 2.95$ \\ \hline
\textbf{DDE-GAN (Ours)} & $\mathbf{0.92 \pm 0.04}$ & $\mathbf{28.12 \pm 1.92}$ \\ 
\hline
\end{tabular}

\end{table}

\section{Conclusions}
In this work we proposed a new methods for synthesis of PET images from CT based on dual domain Generative Adversarial Network. We extend the framework using the property of rotational equivariance which is inherited from the acquisition geometry of both PET and CT imaging. By enforcing this property in the training loss, the proposed DDE-GAN algorithm is able to improve substantially the quality of the PET synthetic images from CT. The results on clinical dataset show an increase of $\approx 2.7$ dB in PSNR and $0.12$ of SSIM compared to the DD-GAN without equivariance. These preliminary observations confirm that DDE-GAN is a potential promising method for image synthesis in clinical applications.  

\section{Acknowledgment}
This work involved human subjects or animals in its research. Approval of all ethical and experimental procedures and protocols was granted by the Research Ethics Committee of McGill University Health Center (Protocol Number: MM-JGH-CR15-50) for the clinical study entitled "HEad and neCK TumOR segmentation and outcome prediction: HECKTOR 2022". A. Altalib is supported by the Jordan University of Science and Technology PhD scholarship. All authors declare that they have no known conflicts of interest in terms of competing financial interests or personal relationships that could have an influence or are relevant to the work reported in this paper.

\bibliographystyle{IEEEtran}
\bibliography{refs}

\end{document}